\documentclass[runningheads]{llncs}
\pdfoutput=1

\usepackage[T1]{fontenc}
\usepackage[dvipsnames]{xcolor}
\usepackage{cite}
\usepackage[hyphens]{url}
\usepackage[accsupp]{axessibility}
\usepackage[labelfont=bf,font=small,tableposition=top]{caption}
\DeclareCaptionFont{tblcap}{\fontsize{6}{6.6}\selectfont}

\usepackage{graphicx}
\usepackage{adjustbox}
\usepackage{booktabs}
\newcommand{\hdrstrut}{\rule{0pt}{2.5ex}}
\usepackage{multirow}
\usepackage{pifont}
\usepackage{amsmath}
\usepackage{amssymb}

\newcommand{\cmark}{\ding{51}}

\usepackage[breaklinks,hidelinks]{hyperref}

\usepackage{orcidlink}
\usepackage{marvosym}

\begin{document}
\pagestyle{empty}

\title{Multi-Scale Latent Temporal Compression Attention and Video-Text Cross Attention Fusion for Video Action Detection}

\titlerunning{MLTC-Attention and VTCA-Fusion for Video Action Detection}

\author{Yicheng Qiu\orcidID{0009-0006-8841-8648} \and
Keiji Yanai\textsuperscript{(\Letter)}\orcidID{0000-0002-0431-183X}}

\authorrunning{Y. Qiu and K. Yanai}

\institute{The University of Electro-Communications, Tokyo, Japan\\
\email{qiu-y@mm.inf.uec.ac.jp, yanai@cs.uec.ac.jp}}

\maketitle

\begin{abstract}
In video action detection, modeling the temporal dependencies in continuous video streams is crucial. Previous methods based on deep neural networks can perform frame-level video recognition tasks. However, these methods struggle to model dynamic temporal patterns, facing challenges such as redundant frame information, difficulty balancing global and local features, and insufficient fusion of visual and textual features. To address these issues, we propose a video temporal attention module based on the Multi-Scale Latent Temporal Compression Attention, which aggregates continuous temporal sequences into feature representations across diverse temporal scales while efficiently reducing redundancy. We also introduce a Video-Text Cross Attention Fusion module that leverages cross-attention at both latent and standard feature levels, achieving fine-grained fusion between video and textual features to enhance semantic alignment and feature interaction. Experimental results across multiple video recognition benchmarks demonstrate that our proposed methods achieve superior performance compared to existing methods, validating the effectiveness and generalization capability.
\end{abstract}

\section{Introduction}
\label{sec:intro}
Video action detection is a critical task in multimodal video understanding, with a wide range of real-world applications such as intelligent surveillance\cite{liu2024memory, qiu2025fire, 10792910}, autonomous driving\cite{chen2020survey, ma2023m2dar}, human behavior analysis\cite{xarles2023astra, qiu2024multipath}, and embodied intelligence\cite{cheng2024videgothink}. Robust models are required to recognize and temporally localize predefined actions as well as abnormal events such as accidents or illegal activities in long, untrimmed video streams. Early approaches extract spatio-temporal features with 3D convolutional networks, while recent methods build on Transformer architectures, state space models\cite{gu2023mamba, qiu2026dst}, and multimodal pre-training to model longer temporal context and to align visual content with category semantics. Although recent Multimodal Large Language Models (MLLMs)\cite{zanella2024harnessing, ye2025vera} generalize well on general video tasks, their limited capacity for long-video temporal reasoning, generative uncertainty, and heavy computational cost make them ill-suited for fine-grained video action detection, motivating specialized architectures with efficient temporal attention and multimodal fusion.
\begin{figure}[t]
  \centering
  \begin{minipage}[b]{0.52\linewidth}
    \centering
    \includegraphics[width=\linewidth]{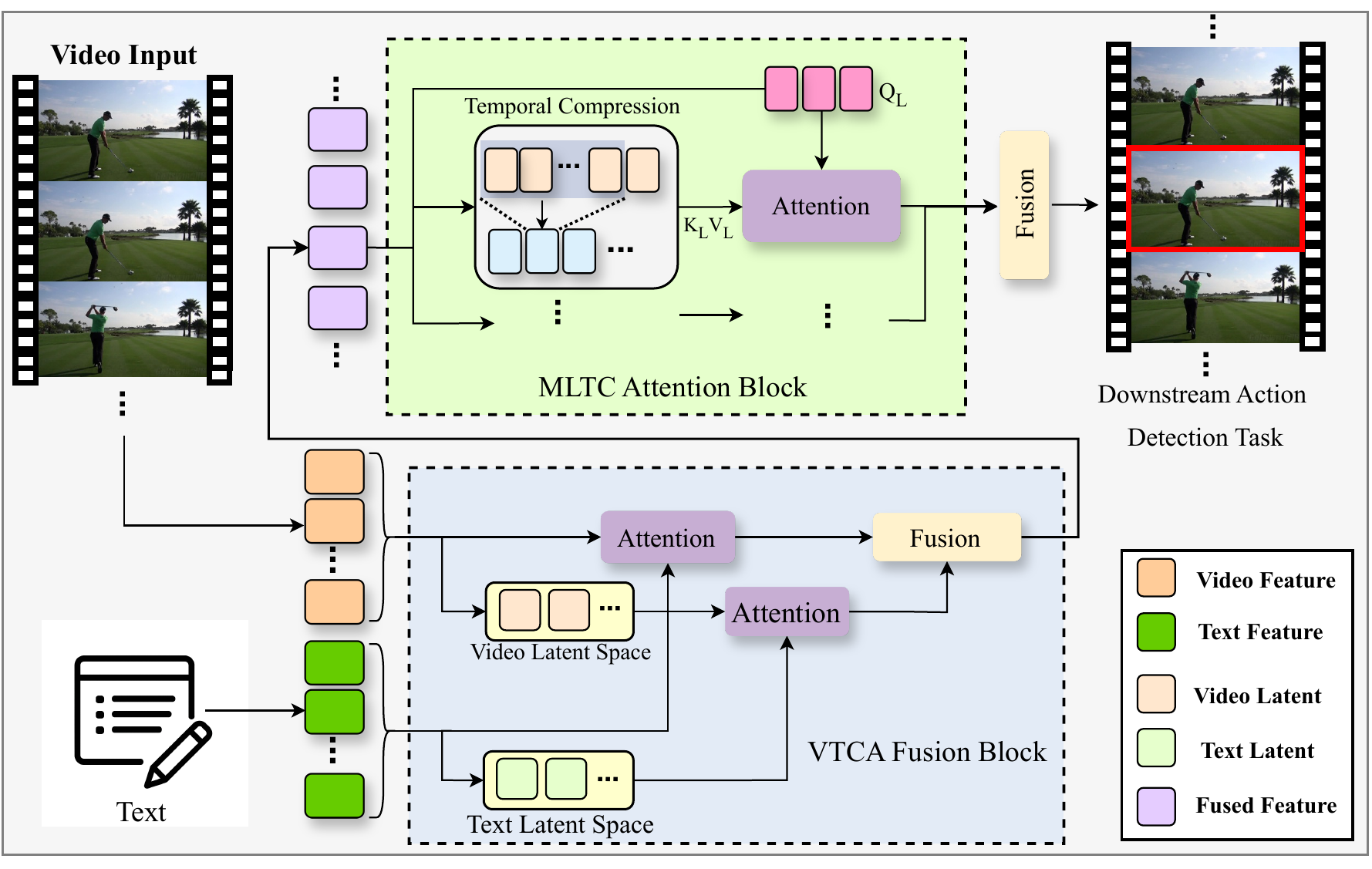}\\[2pt]
    {\scriptsize (a) Framework overview}
  \end{minipage}\hfill
  \begin{minipage}[b]{0.46\linewidth}
    \centering
    \includegraphics[width=\linewidth]{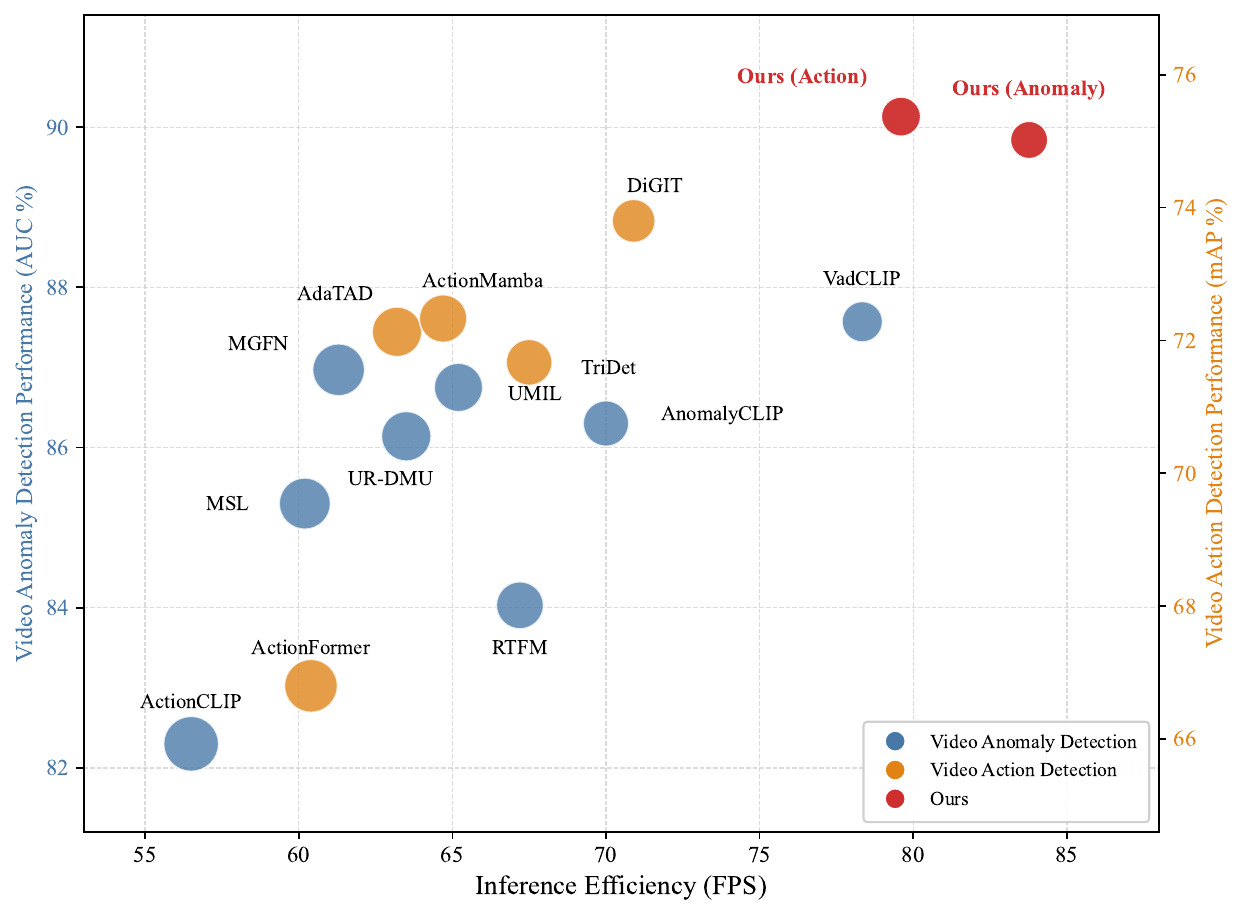}\\[2pt]
    {\scriptsize (b) Performance vs.\ efficiency}
  \end{minipage}
  \captionsetup{font=scriptsize}
  \caption{(a) MLTC-Attention models multi-scale temporal dependencies via compress-then-attend in latent space, and VTCA-Fusion enforces hierarchical video-text alignment at both standard and latent feature levels. (b) Performance--efficiency on video anomaly detection (blue, left axis) and video action detection (orange, right axis): our method (red) attains the best accuracy at the highest inference speed (FPS) with the fewest FLOPs on both tasks.}
  \label{fig:firstpage}
\end{figure}
Transformer attention\cite{vaswani2017attention} is widely used to model temporal context in videos, but two problems remain. First, attention over long sequences repeatedly processes a large amount of redundant frame information, which limits both feature quality and training efficiency. Existing remedies either apply pooling\cite{tang2023temporalmaxer}, which irreversibly discards fine-grained details, or window-based attention\cite{liu2022video}, which restricts the receptive field and misses long-range dependencies; recent efficient temporal models\cite{chen2024video} still struggle to obtain global context and local precision at the same time. To address this issue, we propose a video temporal attention module based on Multi-Scale Latent Temporal Compression Attention, termed MLTC-Attention. The key insight is to first project features into a compact latent space and then compress the temporal dimension at multiple scales before performing attention, rather than attending to all raw features directly. This compress-then-attend design, inspired by recent sparse and compressed attention\cite{wang2020linformer, yuan2025native}, filters temporal redundancy while larger compression scales preserve global context and smaller scales preserve local details, and a gate re-weights the scales according to the input content.

Second, multimodal learning has become a key route to better video understanding, but previous video-text methods\cite{zhu2023languagebind, wang2024multimodal} align video and text features at a single representation level, which limits the depth of cross-modal interaction. To address this limitation, we propose a multimodal feature fusion module, termed Video-Text Cross Attention Fusion (VTCA-Fusion), which enforces cross-modal alignment at two complementary levels: the standard feature space captures explicit semantic correspondences, while a parallel latent space branch, inspired by latent space feature interaction\cite{zanella2024delving}, captures compressed cross-modal patterns. By exploiting the shared category information between video and text, the module achieves fine-grained alignment and complementarity, producing fused features that are more discriminative. As shown in Fig.~\ref{fig:firstpage}, our framework integrates MLTC-Attention and VTCA-Fusion for video action detection. To summarize, the main contributions of this paper are as follows:
\begin{itemize}
\item MLTC-Attention, which compresses features in a multi-scale temporal latent space and adaptively extracts dynamic features across different temporal scales. It strikes a balance between global and local feature extraction, thereby enhancing the ability to efficiently process temporal features in video.

\item VTCA-Fusion, which utilizes cross attention between video features and enhanced category text features at both the standard and the latent level. It overcomes the limitations of existing methods in fusing video and text features, and achieves fine-grained alignment and complementarity between the two modalities.

\item The framework, built around the MLTC-Attention and VTCA-Fusion modules, is applicable to various video action detection tasks. Experimental validation on four benchmarks demonstrates the effectiveness of the proposed modules.

\end{itemize}

\section{Related Work}
\label{sec:relatedwork}
\subsection{Video Understanding Models}
Video representation learning has progressed from 3D CNNs such as C3D\cite{tran2015learning} and I3D\cite{carreira2017quo}, to Transformers such as ViViT\cite{arnab2021vivit} and Video Swin Transformer\cite{liu2022video}, self-supervised VideoMAE\cite{wang2023videomae} and InternVideo\cite{wang2024internvideo2}, and SSM-based VideoMamba\cite{li2025videomamba}. On the multimodal side, CLIP\cite{radford2021learning}, ImageBind\cite{girdhar2023imagebind}, and LanguageBind\cite{zhu2023languagebind} align vision with language and other modalities, Video-LLaVA\cite{lin2023video} and InternVL\cite{chen2024internvl} connect video encoders with large language models, and M2-CLIP\cite{wang2024multimodal} adapts multimodal models to video action recognition.

\subsection{Video Action Detection Models}
Temporal action detection has evolved from boundary-matching methods like BMN\cite{lin2019bmn} and graph-based G-TAD\cite{xu2020g} to Transformer-based approaches such as ActionFormer\cite{zhang2022actionformer}. TriDet\cite{shi2023tridet} employs multi-scale feature pyramids, TemporalMaxer\cite{tang2023temporalmaxer} uses max-pooling, ActionMamba\cite{chen2024video} and efficient temporal attention with state space models\cite{qiu2024efficient, qiu2026efficient} leverage SSM\cite{gu2023mamba} for global temporal modeling, and recent works\cite{li2025temporal, fang2025boundary} explore cross-layer decoupling and boundary refinement.

\subsection{Video Anomaly Detection Models}
MIL-based methods\cite{sultani2018real} learn anomaly scores from video-level labels and were later improved with graph convolutions\cite{zhong2019graph}, feature magnitude learning\cite{tian2021weakly}, multi-sequence Transformers\cite{li2022self}, and dual memory units\cite{zhou2023dual}. Recent methods transfer CLIP\cite{radford2021learning} knowledge to weakly supervised anomaly detection\cite{wu2024vadclip, zanella2024delving, pu2024learning}, use text prompts for normality guidance\cite{yang2024tpwng} or MIL enhancement\cite{chen2024pemil}, and model short and long anomalies across temporal scales\cite{majhi2024oectst}.

\section{Method}

\subsection{Overview}
\begin{figure*}[t]
  \centering
  \captionsetup{font=scriptsize}
  \includegraphics[width=0.90\linewidth]{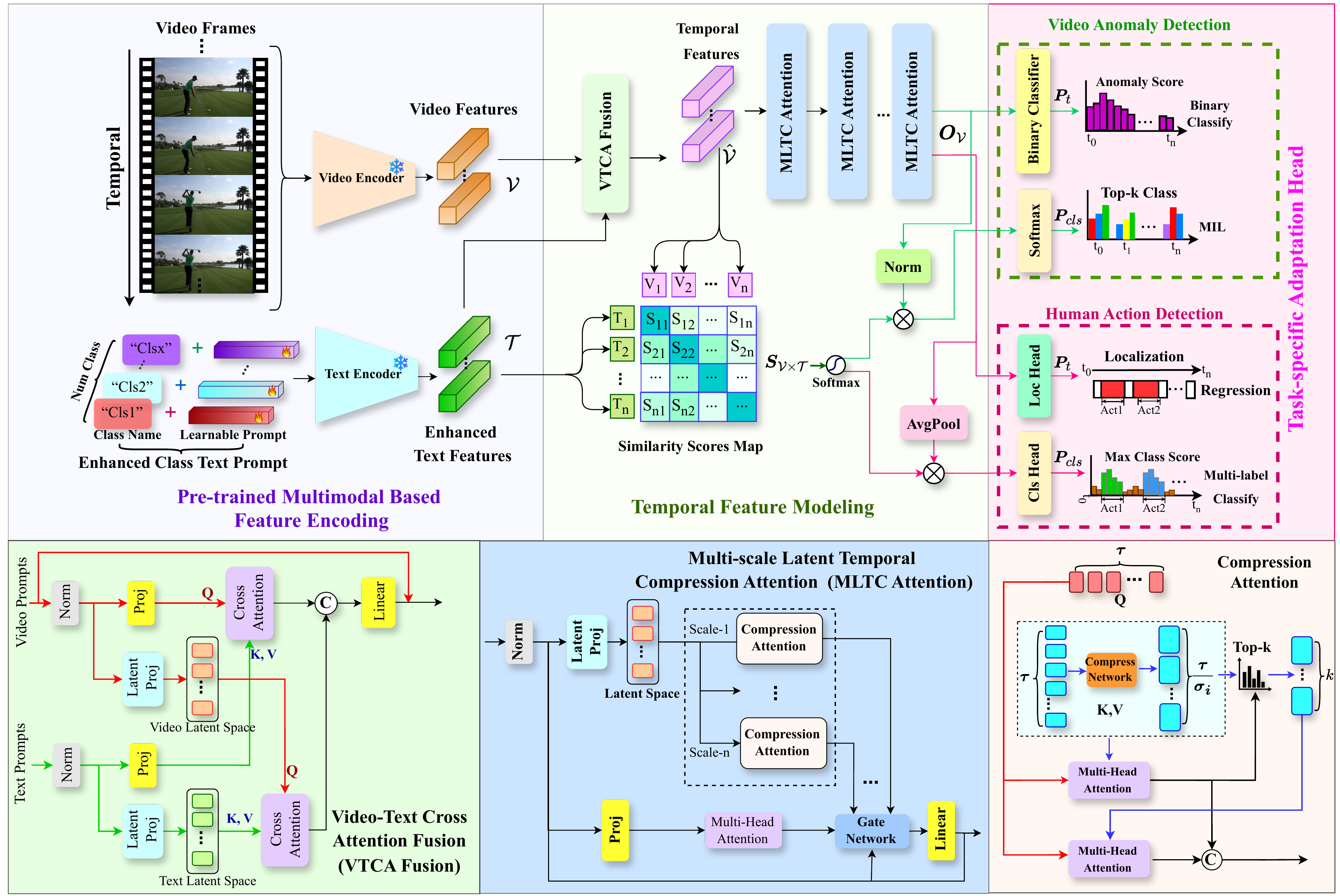}
  \caption{Overview of the overall architecture of our proposed video action detection framework. Our method integrates the VTCA-Fusion module and MLTC-Attention Blocks to process temporal and categorical features, with the resulting predictions feeding into various downstream tasks.}
  \label{fig:overview}
\end{figure*}

Figure~\ref{fig:overview} shows the overall framework. Given a video of $\tau$ frames, the pre-trained video encoder $f_{\mathcal{V}}\left(\cdot\right)$ of the multimodal model extracts frame features $\mathcal{V} = \left[ v_1, \cdots, v_\tau \right] \in \mathbb{R}^{\tau \times d}$. For each of the $c$ categories, the category text $t_c$ is concatenated with the learnable prompt $t_L$, the Enhanced Context, and encoded by the pre-trained text encoder $f_{\mathcal{T}}\left(\cdot\right)$; the resulting text features $\mathcal{T} \in \mathbb{R}^{c \times d}$ are projected into the video feature space. The VTCA-Fusion module fuses $\mathcal{V}$ and $\mathcal{T}$ through standard and latent space cross attention, producing the fused sequence $\hat{\mathcal{V}}=\left[\hat{v}_{1}, \cdots, \hat{v}_{\tau}\right] \in \mathbb{R}^{\tau \times d}$ with the temporal resolution preserved, together with a video-text similarity map $\boldsymbol{S}_{\mathcal{V}\times \mathcal{T}} \in \mathbb{R}^{\tau \times c}$. The MLTC-Attention module then compresses and emphasizes critical temporal information of $\hat{\mathcal{V}}$ in the latent space via multi-scale compression attention, outputting temporal features $\boldsymbol{O}_{\mathcal{V}} \in \mathbb{R}^{\tau \times d}$ that capture both fine-grained details and long-range temporal context. Integrating $\boldsymbol{O}_{\mathcal{V}}$ with $\boldsymbol{S}_{\mathcal{V}\times \mathcal{T}}$ yields the frame-level temporal scores $\boldsymbol{P}_{t} \in \mathbb{R}^{\tau}$ and the class scores $\boldsymbol{P}_{cls} \in \mathbb{R}^{\tau \times c}$, which feed the task-specific loss functions.\par

\subsection{Video-Text Cross Attention Fusion Module}
Standard cross-modal fusion operates at a single representation level, which may overfit to surface-level correlations while missing deeper semantic structures. To overcome this, the VTCA-Fusion module performs cross attention at both standard and latent feature levels simultaneously: the standard-level branch preserves full-dimensional semantic details for explicit alignment, while the latent-level branch operates in a compressed space that encourages the model to learn more abstract cross-modal representations. Its architecture is illustrated in the bottom-left part of Figure~\ref{fig:overview}. We utilize the cross attention, where video frame features attend to text prompt features. Specifically, the video features $\mathcal{V} \in \mathbb{R}^{\tau \times d} $ are treated as queries, while the enhanced text prompt embeddings $\mathcal{T} \in \mathbb{R}^{c \times d} $ serve as both keys and values within a multi-head attention.
We denote $\boldsymbol{Q}_{\mathcal{V}} \in \mathbb{R}^{\tau \times d}, \boldsymbol{K}_{\mathcal{T}} \in \mathbb{R}^{c \times d}, \boldsymbol{V}_{\mathcal{T}} \in \mathbb{R}^{c \times d}$ represent the input for standard multi-head attention,
and $\boldsymbol{Q}^l_{\mathcal{V}} \in \mathbb{R}^{\tau \times d_l}, \boldsymbol{K}^l_{\mathcal{T}} \in \mathbb{R}^{c \times d_l}, \boldsymbol{V}^l_{\mathcal{T}} \in \mathbb{R}^{c \times d_l}$ represent the input for attention of latent space features from the Latent Projection, $d_l < d \in \mathbb{Z^+}$ denotes the output feature dimension in the latent space, which reduces the dimension to compress the feature information, thereby enabling the interaction between video and enhanced text features within the latent space. The attention outputs are computed separately, as follows:
\footnotesize
\begin{equation}\label{eq:standard attention}
  \boldsymbol{Attn}_{\mathcal{V}\!\times\!\mathcal{T}} \!=\! \mathbf{softmax}\!\left(\frac{\boldsymbol{Q}_{\mathcal{V}}\boldsymbol{K}_{\mathcal{T}}^{\top}}{\sqrt{d}} \right)\!\boldsymbol{V}_{\mathcal{T}},
  \quad
  \boldsymbol{Attn}^{l}_{\mathcal{V}\!\times\!\mathcal{T}}\!=\! \mathbf{softmax}\!\left(\frac{\boldsymbol{Q}^{l}_{\mathcal{V}}\boldsymbol{K}^{l^{\top}}_{\mathcal{T}}}{\sqrt{d_l}} \right)\!\boldsymbol{V}^{l}_{\mathcal{T}},
\end{equation}
\normalsize
In Eq.~\ref{eq:standard attention}, $\boldsymbol{Attn}_{\mathcal{V}\times\mathcal{T}} \in \mathbb{R}^{\tau \times d}$ represents the result of standard multi-head attention, where $\boldsymbol{Q}_{\mathcal{V}}\boldsymbol{K}^{\top}_{\mathcal{T}}$ yields the similarity $\boldsymbol{S}_{\mathcal{V}\times\mathcal{T}}$ between the video and enhanced text features, and
$\boldsymbol{Attn}^{l}_{\mathcal{V} \times \mathcal{T}} \in \mathbb{R}^{\tau \times d_l}$ represents the computation result of multi-head attention in latent space.
Finally, we construct an output projection weight tensor $\boldsymbol{W}_{o} \in \mathbb{R}^{(d + d_l) \times d}$ to fuse the outputs of the standard attention and latent space attention, the final result of VTCA-Fusion module is computed, as follows:
\begin{equation}
  \hat{\mathcal{V}} = \mathcal{V} + \boldsymbol{W}_{o}(\boldsymbol{Attn}_{\mathcal{V} \times \mathcal{T}}  \Vert  \boldsymbol{Attn}^{l}_{\mathcal{V} \times \mathcal{T}}).
\end{equation}
where $\Vert$ denotes the operation of concatenating tensors. Through this computation, we obtain the temporal feature $\hat{\mathcal{V}} \in \mathbb{R}^{\tau \times d}$ that fuses the video and text features.\par

\subsection{Multi-Scale Latent Temporal Compression Attention Block}
Standard self-attention\cite{vaswani2017attention} over the full temporal sequence incurs $O(\tau^2)$ complexity, while pooling-based methods\cite{tang2023temporalmaxer} reduce cost but irreversibly discard fine-grained information. Window-based attention\cite{liu2022video} restricts the receptive field and cannot model long-range dependencies. Our MLTC addresses this by first projecting features into a lower-dimensional latent space to reduce per-token cost, then compressing the temporal dimension at multiple scales so that larger scales capture global context while smaller scales preserve local dynamics. Attending to the compressed keys and values reduces the cost of each compression branch from $O(\tau^2)$ to $O(\tau^2/\sigma_i)$, while a single global self-attention branch is retained to preserve the full receptive field.
The architecture of the MLTC is shown in the bottom-center and bottom-right parts of Figure~\ref{fig:overview}. The overall input to the module is the temporal feature $\mathcal{X}_\mathcal{V} \in \mathbb{R}^{\tau \times d}$. First, the Latent Projection maps $\mathcal{X}_\mathcal{V}$ into the latent space, giving $\boldsymbol{Q}^l_t, \boldsymbol{K}^l_t, \boldsymbol{V}^l_t \in \mathbb{R}^{\tau \times \frac{d}{2}}$ as the inputs of the multi-scale latent compression attention. In the bottom-right part of Figure~\ref{fig:overview}, the scales set $[scale\text{-}1, \cdots ,scale\text{-}n]$ denote the compression sizes along the temporal dimension, with each scale corresponding to a Compression Attention module. The total temporal dimension of the Compression Attention input is $\tau$; $\boldsymbol{K}^l_t$ and $\boldsymbol{V}^l_t$ are each transformed along the temporal dimension by their own Compress Network of $scale\text{-}i$, as follows:
\begin{equation}
  \boldsymbol{K}^l_{t_i} = \boldsymbol{W^{out}_{i,K}}\,\mathbf{GELU}(\boldsymbol{W^{in}_{i,K}}\boldsymbol{K}^l_t), \quad
  \boldsymbol{V}^l_{t_i} = \boldsymbol{W^{out}_{i,V}}\,\mathbf{GELU}(\boldsymbol{W^{in}_{i,V}}\boldsymbol{V}^l_t),
\end{equation}
where $\boldsymbol{W^{in}_{i,\ast}} \in \mathbb{R}^{\frac{\tau}{\sigma_i} \times \tau}$, $\boldsymbol{W^{out}_{i,\ast}} \in \mathbb{R}^{\frac{\tau}{\sigma_i} \times \frac{\tau}{\sigma_i}}$, $\ast \in \{K, V\}$, denote the input and output weights of the two Compress Networks, which compress the temporal dimension from $\tau$ to $\frac{\tau}{\sigma_i}$; $i$ indexes the scales set $[scale\text{-}1, \cdots ,scale\text{-}n]$, $\sigma_i \in \mathbb{Z}^+$ is the compression size of $scale\text{-}i$, and $\mathbf{GELU}$ is the activation function. The outputs $\boldsymbol{K}^l_{t_i}, \boldsymbol{V}^l_{t_i} \in \mathbb{R}^{\frac{\tau}{\sigma_i} \times \frac{d}{2}}$ are the inputs of the $i$-th Compression Attention module.
In the Compression Attention module, the computation is first performed based on the input tensors $\boldsymbol{Q}^l_t$ with the compressed $\boldsymbol{K}^l_{t_i}$ and $\boldsymbol{V}^l_{t_i}$ as follows:
\begin{equation}
  \mathcal{S}_i = \mathbf{softmax}\left(\frac{\boldsymbol{Q}^l_t \boldsymbol{K}^{l^\top}_{t_i}}{\sqrt{d/2}} \right), \quad \boldsymbol{Attn}^{cp}_{i} = \mathcal{S}_{i}\boldsymbol{V}^l_{t_i},
\end{equation}
where $\mathcal{S}_{i} \in \mathbb{R}^{\tau \times \frac{\tau}{\sigma_i}}$ represents the similarity scores tensor between $\boldsymbol{Q}^l_t$ and $\boldsymbol{K}^{l}_{t_i}$. $Attn^{cp}_{i} \in \mathbb{R}^{\tau \times \frac{d}{2}}$ represents the computation result of the attention, $cp$ means the compressed $\boldsymbol{K}^{l}_{t_i}, \boldsymbol{V}^l_{t_i}$. Secondly, to capture the critical temporal information from $\boldsymbol{K}^l_{t_i}$ and $\boldsymbol{V}^l_{t_i}$, we perform top-k selection based on $\mathcal{S}_{i}$, as follows:
\begin{equation}
 \mathcal{I}_i = \mathbf{Top\text{-}K} \left(\mathcal{S}_{i}(t), k \right),
\end{equation}
\begin{equation}
  \boldsymbol{K}^{\mathcal{I}_i}_{t_i}\boldsymbol{V}^{\mathcal{I}_i}_{t_i}
  = \big\Vert_{j \in \mathcal{I}_i}
  \boldsymbol{K}^l_{t}\boldsymbol{V}^l_{t}[j\sigma_i:(j+1)\sigma_i],
\end{equation}
where $\mathcal{I}_i \in \mathbb{R}^{k}$ denotes the indices of the top-k most relevant temporal segments in $\boldsymbol{K}^{l}_{t_i}, \boldsymbol{V}^l_{t_i}$ with $\boldsymbol{Q}^l_t$. Following the top-k selection strategy in~\cite{sultani2018real}, we set $k = \max(1, \lfloor \frac{\tau}{\sigma_i} / 8 \rfloor)$, which selects the top $1/8$ of the compressed blocks at each scale with a minimum of 1. The $\boldsymbol{K}^{\mathcal{I}_i}_{t_i} \in \mathbb{R}^{\sigma_i k \times \frac{d}{2}}, \boldsymbol{V}^{\mathcal{I}_i}_{t_i} \in \mathbb{R}^{\sigma_i k \times \frac{d}{2}}$ represent the key and value pairs retrieved via $\mathcal{I}_i$, and they together with $\boldsymbol{Q}^l_t$ , form the input tensors for computing the attention of critical temporal segments, as follows:
\begin{equation}
  \boldsymbol{Attn}^{tk}_{i} = \mathbf{softmax}\left(\frac{\boldsymbol{Q}^l_t \boldsymbol{K}^{\mathcal{I}_i^\top}_{t_i}}{\sqrt{d/2}} \right) \boldsymbol{V}^{\mathcal{I}_i}_{t_i},
\end{equation}
where $Attn^{tk}_{i} \in \mathbb{R}^{\tau \times \frac{d}{2}}$ represents the attention result obtained from the top-k filtering method, enabling the model to enhance feature learning for critical temporal segments.
The results $Attn^{cp}_{i}, Attn^{tk}_{i}$ obtained through the above methods are concatenated, as follows:
\begin{equation}
  \boldsymbol{Attn}_i =  \boldsymbol{Attn}^{cp}_{i} \Vert  \boldsymbol{Attn}^{tk}_{i},
\end{equation}
where $Attn_i \in \mathbb{R}^{\tau \times d}$ is the complete i-th Compression Attention output result. We also apply a Global Attention $Attn_{global}$, which performs standard self-attention \cite{vaswani2017attention} over the entire video sequence $\mathcal{X}_\mathcal{V}$ to capture long-range temporal dependencies and holistic context. Finally, we introduce a gate network with weights applied to each output to fuse Compression Attention across various scales and Global Attention, with the process described as follows:
\begin{equation}
\boldsymbol{O}_\mathcal{V} =  \boldsymbol{W}_{gate} \odot (Attn_{global} \Vert \underbrace{Attn_{1} \Vert \cdots \Vert Attn_{n}}_{n}),
\end{equation}
where
$\boldsymbol{W}_{gate} \in \mathbb{R}^{\tau \times (n+1)}$ denotes the gating weights used to allocate weights to each output, and $\boldsymbol{O}_\mathcal{V} \in \mathbb{R}^{\tau \times d}$ indicates the fused output.  It is further processed by a standard LayerNorm  and FFN  layer to obtain the final representation of the MLTC-Attention module.
Based on temporal feature output $\boldsymbol{O}_{\mathcal{V}}$ and the Video-Text Similarity map $\boldsymbol{S}_{\mathcal{V} \times \mathcal{T}}$, the post-processing step is performed to generate the final temporal and categorical predictions, the $\boldsymbol{P}_{t} \in \mathbb{R}^{\tau}$ represents the temporal prediction scores along the temporal dimension, and the $\boldsymbol{P}_{cls} \in \mathbb{R}^{\tau \times c}$ provides the class-level scores across time.

\subsection{Loss Functions}
\subsubsection{Loss of Video Anomaly Detection.}
Following the standard weakly supervised setting\cite{sultani2018real, zhou2023dual, wu2024vadclip}, only video-level annotations are available for training: given the video-level label $Y \in \{0,1\}$, 0 for normal and 1 for abnormal, we broadcast it along the temporal dimension as $\boldsymbol{y}^i = Y$ for $i = 1,\cdots,\tau$, so that no frame-level annotation is used. A binary cross-entropy loss $\mathcal{L}^{a\text{-}n}_{BCE}$ is computed between the predicted scores $\boldsymbol{P}_{t}$ and $\boldsymbol{y}$. Since an abnormal video still contains normal frames, this broadcast label is inevitably noisy for abnormal videos; the top-$k$ lowest-scoring frames of an abnormal video are therefore additionally selected as normal instances $\boldsymbol{P}_{k\text{-}n} \in \mathbb{R}^k$ with loss $\mathcal{L}^{k\text{-}n}_{BCE}$, which lets the model learn normal features from abnormal videos and mitigates the label noise introduced by the broadcast. Following MIL\cite{sultani2018real}, the mean of the top-$k$ scores of each category in $\boldsymbol{P}_{cls}$ along the temporal dimension gives the predicted logits $\tilde{\boldsymbol{P}}_{cls} \in \mathbb{R}^{c}$, trained with a multi-class cross-entropy $\mathcal{L}^{cls}_{CE}$ against the category labels $\boldsymbol{y}_{cls} \in \mathbb{R}^c$:
\begin{equation}
\footnotesize
\begin{aligned}
\mathcal{L}^{a\text{-}n}_{BCE} &= -\frac{1}{\tau}\sum^{\tau}_{i=1} [\boldsymbol{y}^i \log(\boldsymbol{P}^i_{t}) + (1-\boldsymbol{y}^i)\log(1-\boldsymbol{P}^i_{t})], \quad
\mathcal{L}^{k\text{-}n}_{BCE} = -\frac{1}{k}\sum_{i \in k} \log(1-\boldsymbol{P}^i_{k\text{-}n}), \\
\mathcal{L}^{cls}_{CE} &= -\sum^{c}_{i=1}\boldsymbol{y}^i_{cls}\log(\tilde{\boldsymbol{P}}^i_{cls}), \quad
\mathcal{L}_{a\text{-}n} = \mathcal{L}^{a\text{-}n}_{BCE} + \mathcal{L}^{k\text{-}n}_{BCE} + \mathcal{L}^{cls}_{CE},
\end{aligned}
\end{equation}
where the total loss $\mathcal{L}_{a\text{-}n}$ preserves both the binary anomaly classification and the multi-category loss commonly used in MIL.

\subsubsection{Loss of Temporal Action Detection.}
We use Focal Loss\cite{lin2017focal} for action classification, $\mathcal{L}_{cls}$, and Distance-IoU\cite{zheng2020distance} for temporal segment regression, $\mathcal{L}_{iou}$:
\begin{equation}
  \mathcal{L}_{act} = \frac{1}{\tau^+}\sum^{\tau}_{i=1}[\mathcal{L}_{cls}(\boldsymbol{P}^i_{cls}, \boldsymbol{y}^i_{cls}) + \lambda \mathcal{L}_{iou}(\boldsymbol{P}^i_{t}, \boldsymbol{y}^i_{t}) ],
\end{equation}
where $\boldsymbol{y}_{cls}, \boldsymbol{y}_{t} \in \mathbb{R}^\tau$ are the labels for action categories and temporal segments, $\tau^+$ denotes the number of positive samples assigned during training, and Soft Non-Maximum Suppression is applied as post-processing at inference, and $\lambda$ is the weight of $\mathcal{L}_{iou}$.

\section{Experiments}
\subsection{Datasets and Evaluation Metrics}
For video anomaly detection we use UCF-Crime\cite{sultani2018real}, over 1,900 surveillance videos covering 13 real-world abnormal events together with normal activities, and XD-Violence\cite{wu2020not}, about 4,700 videos from movies and social media covering 6 categories of abnormal events. For temporal action detection we use THUMOS14\cite{THUMOS14}, 413 untrimmed videos with temporal annotations over 20 action categories, and ActivityNet\cite{caba2015activitynet}, 19,994 videos with temporal annotations over 200 action categories. Following common practice, we report the frame-level AUC of the ROC curve on UCF-Crime and the frame-level Average Precision (AP) on XD-Violence, which better reflects the precision--recall balance under class imbalance; for action detection we report the mean Average Precision (mAP) averaged over multiple temporal IoU (tIoU) thresholds between predicted and ground-truth segments.

\subsection{Implementation Details}
We implemented our method on a single NVIDIA L40S GPU. Both pre-trained encoders $f_{\mathcal{V}}(\cdot)$ and $f_{\mathcal{T}}(\cdot)$ are frozen; only the projection layers, the learnable prompts, VTCA-Fusion, MLTC-Attention and the task head are trained, so all text-encoder variants share the same protocol and schedule. During the training process, we use the AdamW\cite{loshchilov2017decoupled} optimizer and employ a scheduler that integrates linear warm-up and cosine annealing to adjust the learning rate, which is set to 1e-5. The batch size was set to 64. For video data input, we utilized pretrained encoder weights to extract video features, and the sequence length of video feature input is 512 frames. Each frame was resized to a resolution of $224 \times 224$ pixels.
For the text input with the enhanced text prompt design, where each textual class label is embedded with 10 learnable prompt sequence length. The maximum text sequence length is set to 79, following the configuration of CLIP\cite{radford2021learning}.
The model was trained for 20 epochs.

\subsection{Evaluation and Comparison}

\begin{table}[t]
  \centering
  \fontsize{6}{6.6}\selectfont
  \captionsetup{font=tblcap}
  \setlength{\tabcolsep}{4pt}\renewcommand{\arraystretch}{0.9}
  \caption{Comparison of frame-level performance between our method and previous methods on the UCF-Crime and XD-Violence benchmarks. $^\dagger$: results reproduced by us with the listed vision encoder.}
  \label{tab:ucf-xd}
  \begin{tabular}{l|c|c|c}
    \toprule
    \multirow{2}{*}{\textbf{Method}} & \multirow{2}{*}{\textbf{Vision Encoder}} & \textbf{UCF-Crime} & \textbf{XD-Violence} \\
    & & AUC(\%) & AP(\%) \\
    \midrule
    LAVAD\cite{zanella2024harnessing} & ImageBind\cite{girdhar2023imagebind} & 80.28 & 62.01 \\
    DMU$^\dagger$\cite{zhou2023dual} & I3D\cite{carreira2017quo} & 86.14 & 81.46 \\
    AnomalyCLIP\cite{zanella2024delving} & ViT-B16\cite{dosovitskiy2020vit} & 86.30 & 77.40 \\
    VERA\cite{ye2025vera} & InternVL2\cite{chen2024internvl} & 86.55 & 70.54 \\
    UMIL\cite{lv2023unbiased} & ViT-B32\cite{dosovitskiy2020vit} & 86.75 & -- \\
    PE-MIL\cite{chen2024pemil} & I3D\cite{carreira2017quo} & 86.83 & -- \\
    MGFN\cite{chen2023mgfn} & VideoSwin\cite{liu2022video} & 86.97 & 79.31 \\
    OE-CTST\cite{majhi2024oectst} & I3D\cite{carreira2017quo} & 86.99 & 81.78 \\
    CLIP-TSA\cite{joo2023clip} & ViT-B16\cite{dosovitskiy2020vit} & -- & 82.15 \\
    EventVAD\cite{shao2025eventvad} & ViT-B16\cite{dosovitskiy2020vit} & 87.51 & 64.04 \\
    VadCLIP$^\dagger$\cite{wu2024vadclip} & I3D\cite{carreira2017quo} & 87.57 & 82.48 \\
    TPWNG\cite{yang2024tpwng} & ViT-B16\cite{dosovitskiy2020vit} & 87.79 & 83.68 \\
    \textbf{MLTC+VTCA+CLIP}\cite{radford2021learning} & I3D\cite{carreira2017quo} & 88.68 & 83.44 \\
    \textbf{MLTC+VTCA+CLIP}\cite{radford2021learning} & ViT-B16\cite{dosovitskiy2020vit} & 88.63 & 84.63 \\
    \textbf{MLTC+VTCA+LangBind}\cite{zhu2023languagebind} & ViT-B16\cite{dosovitskiy2020vit} & 89.17 & 84.89 \\
    \textbf{MLTC+VTCA+DeBERTa}\cite{he2021debertav3} & ViT-B16\cite{dosovitskiy2020vit} & \textbf{89.84} & \textbf{86.53} \\
  \bottomrule
\end{tabular}
\end{table}

\begin{table}[t]
	\centering
	\fontsize{6}{6.6}\selectfont
	\captionsetup{font=tblcap}
	\setlength{\tabcolsep}{3pt}\renewcommand{\arraystretch}{0.9}
	\caption{Comparison on the THUMOS14 and ActivityNet-1.3 with the mAP of tIoU. $^\dagger$: results reproduced by us with the listed vision encoder.}
	\label{tab:thumos}
	\adjustbox{max width=\textwidth}{%
	\begin{tabular}{l|c|cccccc|cccc}
  \toprule
	\multirow{2}{*}{\textbf{Model}} & \multirow{2}{*}{\textbf{Vision Encoder}} &  \multicolumn{6}{c}{THUMOS14} &  \multicolumn{4}{|c}{ActivityNet-1.3} \\
	\cmidrule(lr){3-8}\cmidrule(lr){9-12}
	    &                          & \hdrstrut 0.3     & 0.4    & 0.5     & 0.6  & 0.7     & Avg   & 0.5 & 0.75  & 0.95  &  Avg \\
  \midrule
	ActionFormer\cite{zhang2022actionformer} & I3D\cite{carreira2017quo} & 82.10 & 77.80 & 71.00 & 59.40  & 43.90 & 66.80 & 55.50  & 38.12 & 8.78 & 37.05  \\
  TemporalMaxer\cite{tang2023temporalmaxer}  & InternVidv2\cite{wang2024internvideo2} & 85.23 & 81.27 & 74.17 & 63.70  & 49.14 & 70.70 & 60.99  & 40.53 & 9.79 & 38.59 \\
	TriDet\cite{shi2023tridet}        & InternVidv2\cite{wang2024internvideo2} & 85.81 & 81.78 & 75.61 & 64.64  & 50.50 & 71.67  & 56.40  & 38.74 & 9.60 & 37.60 \\
  Dyfadet\cite{yang2024dyfadet}  & VideoMAEv2\cite{wang2023videomae} & 85.99  & 81.66  & 76.32 & 64.46  &  50.08 & 71.70  & 58.19	 & 39.30  & 8.63  & 38.62  \\
	AdaTAD$^\dagger$\cite{liu2023end}        & InternVidv2\cite{wang2024internvideo2}  & 86.75 & 82.25 & 76.10 & 65.28  & 50.29 & 72.13 & 59.99  & 39.53 & 9.79 & 38.69  \\
  CLTDR-GMG\cite{li2025temporal}  & InternVidv2\cite{wang2024internvideo2}  & 87.00 & 84.00 & 78.40 & 67.90 & 54.00 & 74.30 & 55.00 & 38.00 & 8.60 & 37.10 \\
	ActionMamba\cite{chen2024video}   & InternVidv2\cite{wang2024internvideo2}  & 87.05 & 82.66 & 76.51 & 65.56  & 49.87 & 72.33 & 62.20 & 43.14 & 9.57 & 41.73 \\
  BDRC-Net\cite{fang2025boundary}  & InternVidv2\cite{wang2024internvideo2}  & 88.60 & 83.80 & 78.00 & 66.80 & 51.20 & 73.70 & 61.40 & 42.10 & 10.20 & 41.10 \\
  DiGIT\cite{kim2025digit}  & InternVidv2\cite{wang2024internvideo2}  & 87.60 & 84.20 & 77.60 & 67.30 & 52.50 & 73.80 & 62.00 & 43.10 & 11.30 & 42.00 \\
\textbf{MLTC+VTCA+CLIP}\cite{radford2021learning} & InternVidv2\cite{wang2024internvideo2}
& 88.95 & 85.12 & 78.40 & 68.02  & 52.51 & 74.60 & 63.81 & 45.34 & 10.92 & 43.59 \\
\textbf{MLTC+VTCA+LangBind}\cite{zhu2023languagebind} & InternVidv2\cite{wang2024internvideo2}
& 88.99 & 85.22 & 78.73 & 68.30 & 52.78 & 74.80
& 63.84 & 45.35 & 11.17 & 43.82 \\
  \textbf{MLTC+VTCA+DeBERTa}\cite{he2021debertav3}  & InternVidv2\cite{wang2024internvideo2} & \textbf{89.50} & \textbf{85.38} & \textbf{78.82} & \textbf{69.05}  & \textbf{54.15} & \textbf{75.37}  & \textbf{63.87} & \textbf{45.43} & \textbf{11.36}   & \textbf{43.90} \\
	\bottomrule
	\end{tabular}%
	}
\end{table}

\begin{table}[t]
\centering
\fontsize{6}{6.6}\selectfont
\captionsetup{font=tblcap}
\setlength{\tabcolsep}{3pt}
\begin{minipage}[t]{0.56\textwidth}
  \centering
  \caption{Ablation for module components and for the latent projection in each module.}
  \label{tab:ablation removing}
  \adjustbox{max width=\linewidth}{%
  \begin{tabular}{ccccc|c|c}
    \toprule
    \multicolumn{5}{c|}{Module} & UCF-C & THUMOS \\
    \hline
    MLTC & Top-k & VTCA & Latent (MLTC) & Latent (VTCA) & AUC & mAP \\
    \midrule
     &  & \cmark &        & \cmark & 86.00 & 72.53 \\
    \cmark &  & \cmark & \cmark & \cmark & 88.27 & 73.75 \\
    \cmark & \cmark &  & \cmark &        & 86.65 & 73.23 \\
    \cmark & \cmark & \cmark &  &  & 88.19 & 73.68 \\
    \cmark & \cmark & \cmark & \cmark &  & 89.09 & 74.92 \\
    \cmark & \cmark & \cmark &  & \cmark & 88.92 & 74.66 \\
    \cmark & \cmark & \cmark & \cmark & \cmark & 89.84 & 75.37 \\
    \bottomrule
  \end{tabular}}
\end{minipage}\hfill
\begin{minipage}[t]{0.40\textwidth}
  \centering
  \caption{Ablation results for alternative module design.}
  \label{tab:ablation alter}
  \adjustbox{max width=\linewidth}{%
  \begin{tabular}{ll|c|c}
    \toprule
    \multicolumn{2}{c|}{Attention Modules} & UCF-C & THUMOS \\
    \hline
         Temporal Attn       &   V\&T Fusion     & AUC   &   mAP   \\
    \midrule
    Self-Attn\cite{vaswani2017attention} & VTCA   & 86.00 & 72.53 \\
    Soft-Attn\cite{yan2018hierarchical} & VTCA   & 86.14 & 72.64 \\
    MS-TCT\cite{dai2022ms} & VTCA     & 88.01 & 73.44 \\
    MLTC & Cross-Attn\cite{ouenniche2023vision}  & 88.19 & 73.23 \\
    MLTC & VTCA & 89.84 & 75.37 \\
    \bottomrule
  \end{tabular}}
\end{minipage}
\end{table}

\begin{table}[t]
\centering
\fontsize{6}{6.6}\selectfont
\captionsetup{font=tblcap}
\setlength{\tabcolsep}{3pt}
\begin{minipage}[t]{0.30\textwidth}
  \centering
  \caption{Ablation for different temporal compression scales.}
  \label{tab:ablation scales}
  \adjustbox{max width=\linewidth}{%
  \begin{tabular}{l|c|c}
    \toprule
    \multirow{2}{*}{Scales} & UCF-C & THUMOS \\
    \cmidrule(lr){2-3}
              & \hdrstrut AUC   &   mAP   \\
    \midrule
    w/o   & 86.00 & 72.53 \\
    {[4,16,32]}   & 88.77 & 74.32  \\
   {[8,16,64]}   & 89.05   & 74.57 \\
   {[8,32,64]} & 89.84 & 75.37 \\
   {[16,64,128]} & 88.91 & 74.65 \\
    \bottomrule
  \end{tabular}}
\end{minipage}\hfill
\begin{minipage}[t]{0.32\textwidth}
  \centering
  \caption{Ablation for text prompt types.}
  \label{tab:ablation_prompts}
  \adjustbox{max width=\linewidth}{%
  \begin{tabular}{l|c|c}
    \toprule
    \multirow{2}{*}{Prompt Types} & UCF-C & THUMOS \\
    \cmidrule(lr){2-3}
              & \hdrstrut AUC   &   mAP   \\
    \midrule
    w/o prompt & 86.65 &  73.23\\
    Random  & 86.72 &   73.31  \\
    Generic  & 86.84 &  73.45   \\
    Class-Only  & 88.74     &   74.62  \\
    Learnable   & 89.84 & 75.37 \\
    \bottomrule
  \end{tabular}}
\end{minipage}\hfill
\begin{minipage}[t]{0.34\textwidth}
  \centering
  \caption{Efficiency comparison with prior methods.}
  \label{tab:efficiency}
  \adjustbox{max width=\linewidth}{%
  \begin{tabular}{l|c|c}
    \toprule
    Method &  FPS & FLOPs  \\
    \midrule
    MGFN\cite{chen2023mgfn}   & 61.30 & 57.46G  \\
    AnomCLIP\cite{zanella2024delving}   & 69.98  &  50.4G \\
    STPrompt\cite{wu2024weakly}   & 79.19  & 49.6G \\
    VadCLIP\cite{wu2024vadclip}   & 78.34  & 44.8G \\
    Ours   & 83.77  & 42.1G  \\
    \bottomrule
  \end{tabular}}
\end{minipage}
\end{table}

\noindent \textbf{Results of Video Anomaly Detection.}
Table~\ref{tab:ucf-xd} compares our method with previous state-of-the-art methods on UCF-Crime and XD-Violence. Our method achieves the highest AUC on UCF-Crime and, without incorporating any audio features, the highest AP on XD-Violence. Compared with recent LLM-based approaches, LAVAD\cite{zanella2024harnessing}, EventVAD\cite{shao2025eventvad}, and VERA\cite{ye2025vera}, our method obtains clearly higher accuracy on both datasets without relying on large language models, indicating that LLM-based approaches show no clear advantage on fine-grained temporal localization while requiring substantially higher computational cost, whereas our targeted attention and fusion designs are more effective for fine-grained temporal anomaly detection.

\noindent \textbf{Results of Temporal Action Detection.}
Table~\ref{tab:thumos} reports the results on THUMOS14 and ActivityNet-1.3. On THUMOS14, our method achieves the best average mAP over tIoU thresholds from 0.3 to 0.7 and consistently maintains the best mAP across all thresholds; on ActivityNet-1.3, it also achieves the best overall performance and continues to lead at every tIoU threshold. Our method consistently outperforms both earlier Transformer- and SSM-based detectors and the most recent state-of-the-art methods on both datasets.\par

\noindent \textbf{Computational Efficiency Analysis.}
Table~\ref{tab:efficiency} compares the computational cost and inference speed of our MLTC module with previous methods under the same input resolution and hardware settings, on a single NVIDIA L40S GPU with 512-frame inputs at $224\times224$ resolution, excluding I/O and preprocessing overhead. Our method achieves the highest throughput with the fewest FLOPs among all compared methods while also being the most accurate, reducing the computational cost compared with prior methods while simultaneously improving accuracy. By projecting features into a compact latent space and compressing temporal redundancy before the attention computation, our compress-then-attend paradigm achieves a better accuracy--efficiency trade-off than prior methods that increase complexity for marginal gains.

\subsection{Qualitative Analyses}

\begin{figure}[t]
  \centering
  \captionsetup{font=scriptsize}
  \includegraphics[width=0.75\linewidth]{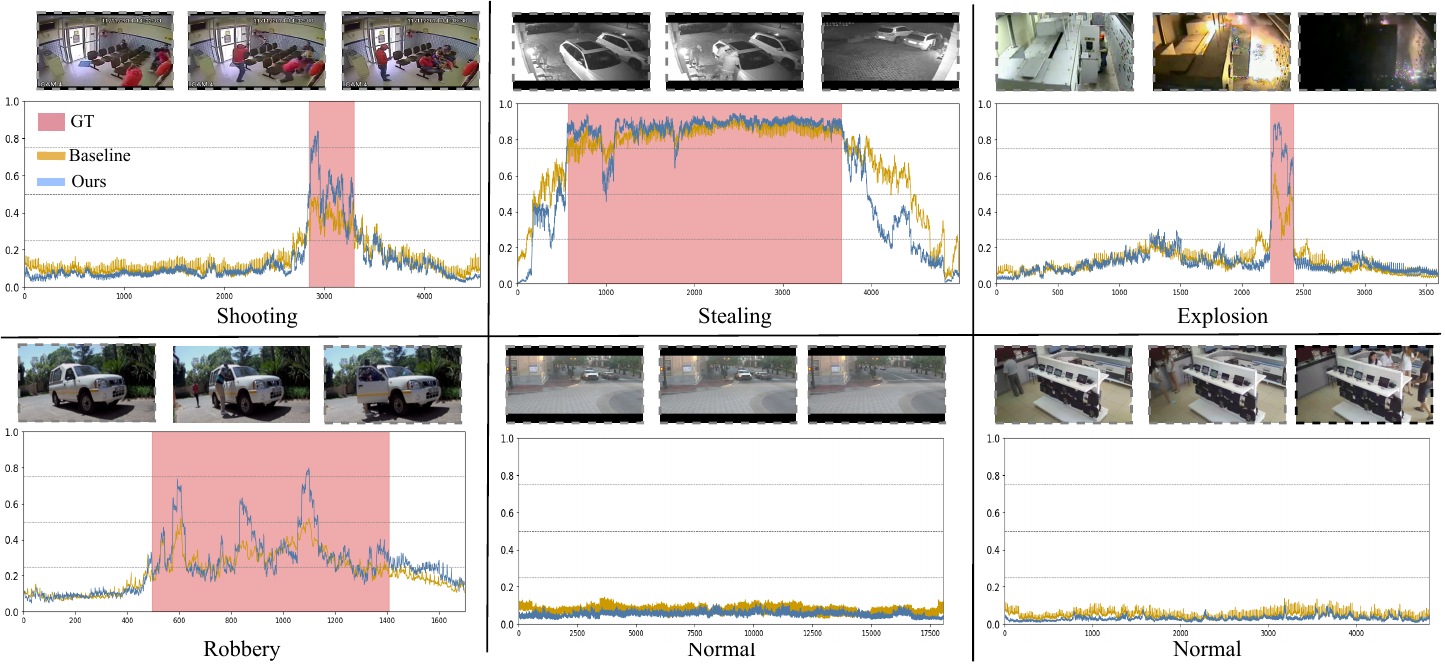}
  \caption{Qualitative results of our proposed methods on Video Anomaly Detection tasks.}
  \label{fig:Qualitative results VAD}
\end{figure}
\begin{figure}[t]
  \centering
  \captionsetup{font=scriptsize}
  \includegraphics[width=0.75\linewidth]{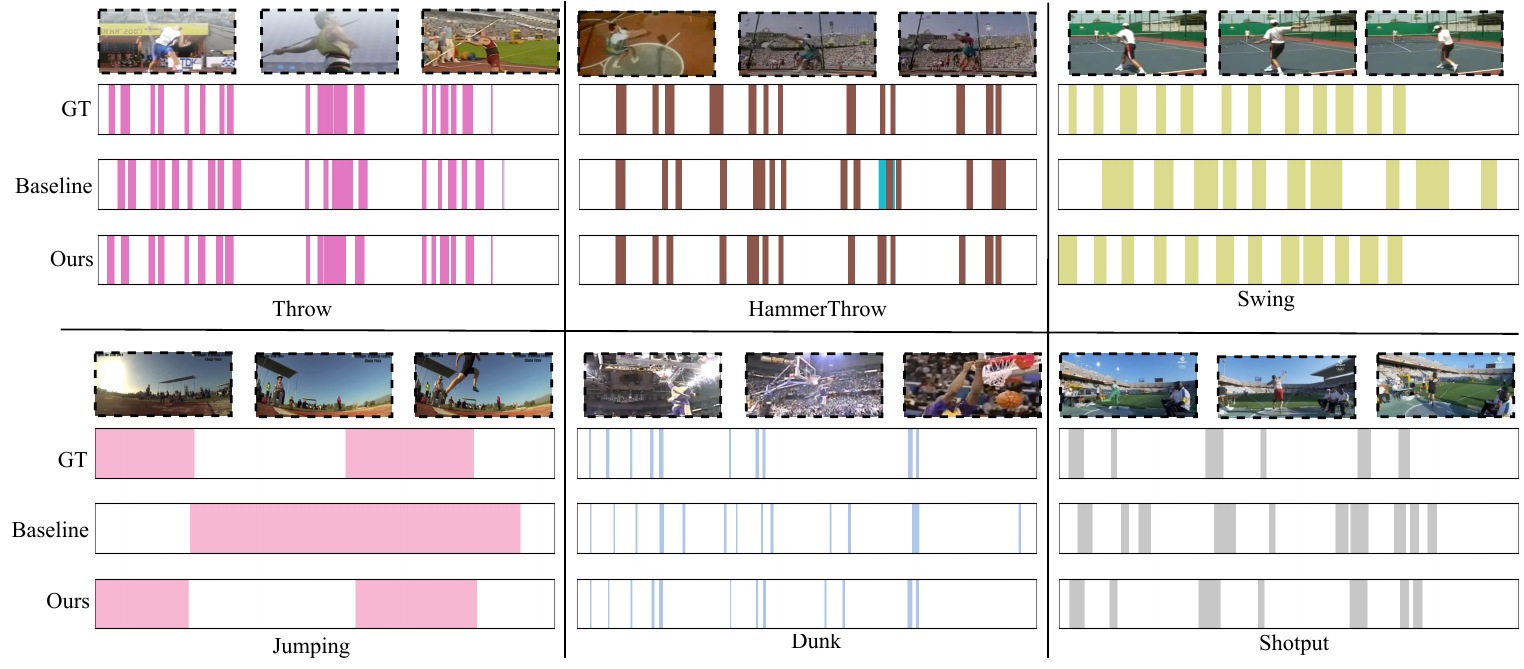}
  \caption{Qualitative results of our proposed methods on temporal action detection tasks.}
  \label{fig:Qualitative results TAL}
\end{figure}
\begin{figure}[t]
  \centering
  \captionsetup{font=scriptsize}
  \includegraphics[width=0.55\linewidth]{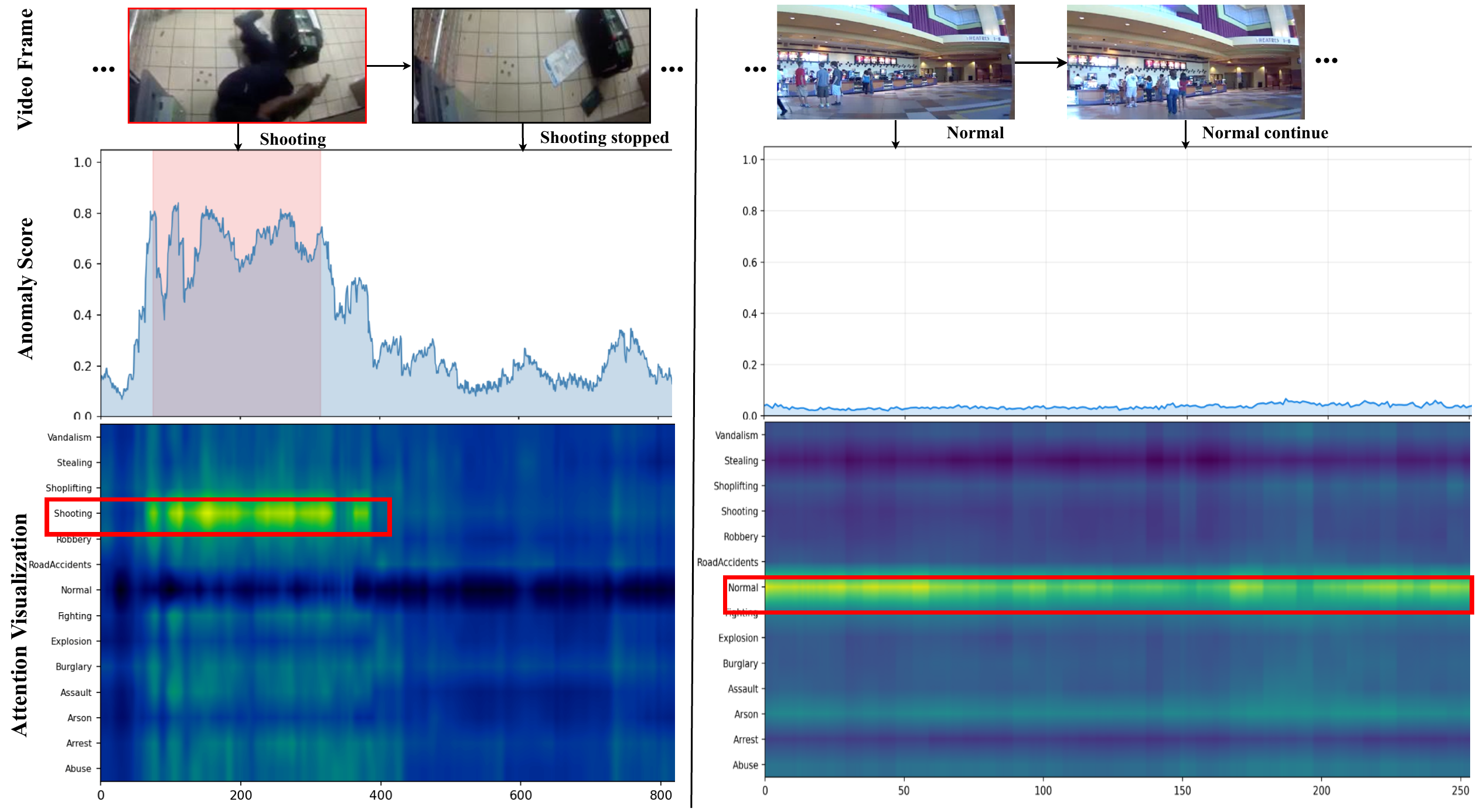}
  \caption{Visualization of internal representation category-wise temporal feature.}
  \label{fig:Qualitative feature}
\end{figure}

We present visualization examples of temporal predictions produced by our model. Figure~\ref{fig:Qualitative results VAD} shows Video Anomaly Detection results, where the curves are predicted anomaly scores over time and the red shaded regions are ground-truth (GT) anomaly intervals, and Figure~\ref{fig:Qualitative results TAL} shows temporal action detection results, where colored segments are predicted action categories and temporal intervals; compared with the baseline, our method yields more precise temporal predictions on both tasks. Figure~\ref{fig:Qualitative feature} visualizes the category-wise temporal attention: for anomaly videos the model correctly activates on the corresponding category during the GT interval, while for normal videos ``Normal'' dominates the response, reflecting the joint effect of MLTC-Attention, which captures fine-grained dynamics across multiple scales, and VTCA-Fusion, which aligns video features with category-specific textual semantics.

\subsection{Ablation Studies}

To better analyze the contribution of each component in our model, we conduct ablation studies on the representative benchmarks UCF-Crime\cite{sultani2018real} and THUMOS14\cite{THUMOS14}.\par
\noindent \textbf{Contribution of Our Proposed Modules.}
As shown in Table~\ref{tab:ablation removing}, we ablate each module within the overall architecture and evaluate on representative datasets for each task. First, for the MLTC-Attention module, we replace it with standard attention. The results on both UCF-Crime and THUMOS14 datasets show a noticeable performance decline, demonstrating the superiority of MLTC-Attention in capturing temporal features.  We also conducted experiments on the effect of the Top-k selection, which retains the top 1/8 of the compressed blocks at each scale, at least one block, following strategies in previous works\cite{shi2023tridet}. Its removal led to a slight performance decrease on both datasets, further validating the effectiveness of this filtering mechanism in focusing on the most informative temporal segments. When the VTCA-Fusion module is removed, the correlation between video and text features weakens, resulting in a performance drop on both tasks. Finally, the latent component, which applies latent space attention within both the MLTC-Attention and VTCA-Fusion modules, is crucial for understanding key semantic features in latent space; its removal degrades performance significantly.\par

\noindent \textbf{Alternative of Our Proposed Modules.}
We replace MLTC-Attention or VTCA-Fusion with several previously used temporal and cross-modal attention modules. As shown in Table~\ref{tab:ablation alter}, substituting MLTC with Self-Attention, Soft-Attention, or MS-TCT consistently lowers performance, indicating that multi-scale temporal compression is crucial for modeling long-range dependencies. Similarly, replacing VTCA with standard Cross-Attention also degrades results, demonstrating that our multi-level video-text fusion achieves more effective cross-modal alignment. \par

\noindent \textbf{Multi-scale Compression factors.}
To analyze the effect of different temporal compression scales on the MLTC-Attention module, we conducted an ablation study by varying scales. As shown in the Table~\ref{tab:ablation scales}, the $w/o$ denotes the absence of compression scales, corresponding to the standard attention. The smaller compression set [4, 16, 32] preserves fine-grained temporal details but introduces redundant computations, while the larger compression set [16, 64, 128] overly compresses the temporal dimension, leading to the loss of local dynamic characteristics. The configuration [8, 32, 64] achieves a better balance between feature compression and redundancy reduction. Intuitively, the larger scales 32 and 64 capture global anomaly regions across the temporal extent, while the smaller scale 8 refines local action boundaries, and their combination achieves sharper activation within ground-truth intervals.\par

\noindent \textbf{Prompt Types.}
As shown in Table~\ref{tab:ablation_prompts}, removing textual input, denoted $w/o$ in the table, clearly degrades performance, indicating that text provides useful semantic priors. Random and generic prompts introduce only weak guidance, as they lack category-aligned semantics. Using Class-Only prompts brings some class information but remains limited. In contrast, the Learnable prompt achieves the best results, showing that learnable prompt tokens effectively capture category-specific semantics and significantly improve multimodal feature learning.

\section{Conclusion}
We tackle key problems in multimodal video action detection by our novel solutions. First, we propose the MLTC-Attention module, which utilizes a multi-scale temporal latent space compression attention. It is designed to adaptively capture important dynamic features across different temporal scales, effectively reducing the computational redundancy and inefficiency. Second, we present the VTCA-Fusion module, which employs cross attention between video content and enhanced multi-category text. This enables precise alignment and complementary integration of visual and textual features, addressing limitations in existing multimodal feature integration methods.
Extensive experiments on temporal action detection and video anomaly detection benchmarks demonstrate that our methods consistently improve accuracy and achieve superior overall performance.

\begin{credits}
\subsubsection{\ackname} This work was supported by JST CRONOS JPMJCS24K4 and JSPS KAKENHI 26K23875.

\subsubsection{\discintname} The authors declare no competing interests.
\end{credits}

\bibliographystyle{splncs04}
{\def\small{\fontsize{7.5}{8.2}\selectfont}
\bibliography{main}}

\end{document}